\documentclass{article}
\usepackage[utf8]{inputenc}

\usepackage{boxinz-macros}
\usepackage{svg}
\usepackage{xcolor}
\usepackage{todonotes}
\usepackage[T1]{fontenc}

\usepackage{authblk}

\providecommand{\keywords}[1]
{
  \small	
  \textbf{\textit{Keywords---}} #1
}

\title{Personalized Binomial DAGs Learning with Network Structured Covariates}

\author[1]{Boxin Zhao$^*$}
\author[2]{Weishi Wang$^*$}
\author[3]{Dingyuan Zhu}
\author[3]{Ziqi Liu}
\author[3]{Dong Wang}
\author[3]{Zhiqiang Zhang}
\author[3]{Jun Zhou}
\author[4]{Mladen Kolar}
\affil[1]{\textit{Booth School of Business, University of Chicago}}
\affil[2]{\textit{Industrial and Operations Engineering, University of Michigan}}
\affil[3]{\textit{Ant Financial Group}}
\affil[4]{\textit{Department of Data Sciences and Operations, Marshall School of Business, University of Southern California}}

\date{}

\begin{document}

\maketitle

\def\thefootnote{*}\footnotetext{Equal contributions.}

\begin{abstract}
\normalsize
The causal dependence in data is often characterized by Directed Acyclic Graphical (DAG) models, widely used in many areas. Causal discovery aims to recover the DAG structure using observational data. This paper focuses on causal discovery with multi-variate count data. We are motivated by real-world web visit data, recording individual user visits to multiple websites. Building a causal diagram can help understand user behavior in transitioning between websites, inspiring operational strategy. A challenge in modeling is user heterogeneity, as users with different backgrounds exhibit varied behaviors. Additionally, social network connections can result in similar behaviors among friends. We introduce personalized Binomial DAG models to address heterogeneity and network dependency between observations, which are common in real-world applications. To learn the proposed DAG model, we develop an algorithm that embeds the network structure into a dimension-reduced covariate, learns each node's neighborhood to reduce the DAG search space, and explores the variance-mean relation to determine the ordering. Simulations show our algorithm outperforms state-of-the-art competitors in heterogeneous data. We demonstrate its practical usefulness on a real-world web visit dataset.
\end{abstract}


\section{Introduction}
Probabilistic directed acyclic graphical (DAG) models provide a powerful tool for modeling causal or directional dependence relationship among multiple variables, with applications in various domains~\citep{doya2007bayesian, friedman2000using, jansen2003bayesian, kephart1992directed}. However, the underlying DAG often needs to be learned from observational data. Causal discovery or causal structure learning research addresses this challenge.

We address causal discovery in multi-variate count data. Let $\mathbf{X}=(X_1,\ldots,X_{d_X})^{\top}$ be a random vector, where each $X_j \in \{0,1,\ldots,T\}$ is a count on the $j$-th record, with $T$ as the maximum. Our goal is to uncover the DAG structure that factorizes the distribution of $\mathbf{X}$. Web visit data serves as a motivating example, where each observation corresponds to the number of visits by a user to $d_X$ websites in the last $T$ records -- $X_j$ is the count of visits by a user to the $j$-th website. We aim to understand directional relationships between websites. Specifically, if users visit website $k$ after website $j$, there is a directional relationship $j \to k$. Modeling these relationships with DAGs reveals user transition logic, aiding operational strategy design.

Existing research on causal discovery with such data type~\citep{park2017learning} assumes identically independent distributed observations, which is often not the case. For example, in web visit data, users may share a single DAG structure but have edge weights (transition probabilities) dependent on features like age, sex, and consumption habits. Additionally, socially connected users tend to exhibit similar behaviors. These two concerns motivate us to propose personalized Binomial DAG models to address heterogeneity and social network structure of the observations. Specifically, we allow each observation to have a personalized DAG model. We assume that different observations share the same DAG structure but have weights dependent on embedded features and network structure. Our model offers a novel approach to causal discovery with dependent observations.

We propose a learning algorithm to recover the DAG structure from observations. First, we embed the observation feature vector into a low-dimensional covariate vector, considering the topology of the observations' network. We suggest two options: eigen-decomposition for linear embedding~\citep{zhao2022dimension} and Graph Auto-Encoders (GAEs) for nonlinear embedding~\citep{kipf2016variational}. Next, we learn the neighborhood set of each node. To allow DAG weights to depend on embedded features, we use penalized kernel smoothing~\citep{kolar2010sparse}, encouraging similar embedded features to have similar DAG weights while maintaining the same DAG structure. After obtaining the neighborhood set, we decide the DAG ordering using an overdispersion score, which utilizes the relationship between the mean and variance~\citep{park2017learning}. Finally, we repeat the second step to obtain the final DAG structure.

We demonstrate our method with experiments on simulated and real-world data. In simulations, we compare it with the state-of-the-art method~\citep{park2017learning} that ignores observation heterogeneity. When data generation is heterogeneous, the competitor method performs poorly, but ours achieves consistent DAG structure recovery as sample size increases. For real-world data, our method applied to web visit data reveals insights into Chinese customers' behavior during the COVID-19 pandemic.

\textbf{Contributions:} Our paper makes the following contributions.
\begin{itemize}

\item We introduce personalized Binomial DAG models, incorporating observation features and social network structure, to study causal discovery with dependent observations.

\item We propose an efficient algorithm to recover the DAG structure from observations.

\item Simulation experiments show our method outperforms the state-of-the-art competitor in heterogeneous data scenarios.

\item We demonstrate the practical usefulness of our method on a web visit dataset, revealing insights into Chinese customers' behavior during COVID-19.

\end{itemize}

 
\textbf{Organization:} We summarize related work in Section~\ref{sec:related-work}. Section~\ref{sec:prelim} defines the problem setup and data generation assumptions. Section~\ref{sec:methodolgy} presents our four-step DAG learning algorithm. Section~\ref{sec:hyperparameter} covers hyperparameter tuning. Section~\ref{sec:simul-exp} validates our algorithm with simulations. Finally, Section~\ref{sec:real-data-exp} applies our algorithm to web visit data.
The code of the paper is available at \url{https://github.com/DiegoWangSys/PersonalizedDAG}.

\section{Related Work}
\label{sec:related-work}

This section summarizes related work. Our method combines node embedding, penalized kernel smoothing, and overdispersion.

The first step of our algorithm encodes network structure and observation features into a low-dimensional covariate vector, known as node or network embedding in graph learning. Graph Neural Networks (GNNs) are powerful tools for this~\citep{wu2020comprehensive}. Deep learning methods build an encoder to embed node features into a low-dimensional latent space and a decoder to recover graph topology. Some methods do not require node features~\citep{cao2016deep,wang2016structural}, while others use node features~\citep{pan2018adversarially}, or both node and edge features~\citep{li2018learning,simonovsky2018graphvae}. We apply Graph Auto-Encoder (GAE)~\citep{kipf2016variational} for network embedding with or without node features. Besides deep learning methods, non-deep learning methods include matrix factorization~\citep{shen2018discrete,yang2018binarized} and random walks~\citep{perozzi2014deepwalk}. Recently, \citet{zhao2022dimension} proposed a linear node embedding method with good theoretical properties under linear assumptions. We recommend this method for linear node embedding.


Kernel smoothing is a popular nonparametric method to estimate real-valued functions, particularly probability density functions~\citep{wand1994kernel}. \citet{kolar2010sparse} proposes a penalized kernel smoothing estimator for non-zero elements of the precision matrix. We adopt a similar approach to allow personalized edge weights.

Overdispersion is a characteristic of random variables where the variance is directly proportional to the mean. It occurs when the observed variance is greater than the variance predicted by a theoretical model, such as the Generalized Linear Mixed Models (GLMM)~\citep{aitkin1996general}. This phenomenon has been used in many applications~\citep{dean1992testing,ravikumar2010high}. \citet{park2015learning,park2017learning} use overdispersion to address model identifiability. We also apply overdispersion to determine the ordering of our binomial DAG model.

\section{Preliminaries}
\label{sec:prelim}

Consider a data set $\mathcal{D}^n = \{ \mathbf{X}^{(i)}, \mathbf{Z}^{(i)} \}_{i=1}^n$, where $\mathbf{X}^{(i)} = (X_1^{(i)}, \ldots, X_{d_X}^{(i)})^\top \in \mathcal{X}$, with $\mathcal{X} = \{0, 1, \ldots, \allowbreak T\}^{d_X}  \subseteq \mathbb{N}^{d_X}$, and $\mathbf{Z}^{(i)} = (Z_1^{(i)}, \ldots, Z_{d_Z}^{(i)})^\top \in \mathcal{Z} \subseteq \mathbb{R}^{d_Z}$. Here, $\mathbf{X}^{(i)}$ is a random vector whose distribution we investigate using a DAG, while $\mathbf{Z}^{(i)}$ are covariates that aid in the inference of the DAG structure for $\mathbf{X}^{(i)}$. In our application, $X_j^{(i)}$ denotes the number of visits to the website $j$ by the user $i$ over the last $T$ records. The vector $\mathbf{Z}^{(i)}$ includes features such as age, sex, and profession of the user $i$. Thus, the dataset integrates the observed count data with the corresponding covariates.

We assume that there is a known relationship network associated with observations. We represent this network with an undirected graph $G=(V,E)$, where $V=[n]$ is the node set and $E \subseteq V \times V$ is the edge set. Each node of $V$ represents an observation, and an edge $(i,j) \in E$ indicates a connection between observation $i$ and $j$. In our motivating example, connections represent the contact or payment history between users. Connected observations often have similar distributions, which makes network information useful for learning the DAG structure.

Let $\mathcal{D}^n_Z=\{\mathbf{Z}^{(i)}\}^n_{i=1}$ be deterministic. We explore the DAG structure of $\mathbb{P} (\mathbf{X}^{(i)} \mid \mathcal{D}^n_Z, G )$ for $i=1,\ldots,n$. Let $G_X=(V_X, E_X)$ be a directed graph without cycles (DAG) with vertex set $V_X=[d_X]$ and edge set $E_X \subseteq V_X \times V_X$. A directed edge from node $j$ to $k$ is denoted by $(j,k) \in E_X$ or $j\to k$. Node $j$ is the parent of node $k$, and node $k$ is the child of node $j$. Let $\textrm{pa}(j)$ be the set of all parents of node $j$, and $\textrm{ch}(j)$ be the set of all children of node $j$. If there is a directed path $j \to \ldots \to k$, then $j$ is an ancestor of $k$ and $k$ is a descendant of $j$. Let $\textrm{de}(j)$ be the set of descendants of $j$ and $\textrm{an}(j)$ be the set of ancestors of $j$. Define $j \in \textrm{de}(j)$ and $j \in \textrm{an}(j)$. Node $j$ is a root node if $\textrm{pa}(j)=\emptyset$. By the definition of DAG, there is at least one root node.

Our first assumption states that a single DAG factorizes all $\mathbb{P} (\mathbf{X}^{(i)} \mid \mathcal{D}^n_Z, G )$.
\begin{assump}[Common DAG Structure]
\label{assump:comm-dag}
There exists a DAG $G_X=(V_X,E_X)$ such that for all $i=1,\ldots,n$, we have the following factorization:
\begin{equation*}
\mathbb{P} \left( \mathbf{X}^{(i)} \mid \mathcal{D}^n_Z, G \right) = \prod_{j \in V_X} f_j \left( X^{(i)}_j \mid \mathcal{D}^n_Z, \mathbf{X}^{(i)}_{ \textrm{pa}(j) }, G \right),
\end{equation*}
where $f_j ( X^{(i)}_j \mid \mathcal{D}^n_Z, \mathbf{X}^{(i)}_{ \textrm{pa}(j) }, G )$ represents the conditional distribution of $X^{(i)}_j$ given $\mathcal{D}^n_Z$, network $G$ and its parent variables $X^{(i)}_k, k \in \textrm{pa}(j)$.
In addition, we assume that $\mathbf{X}^{(i)} \independent \mathbf{X}^{(l)} \mid \mathcal{D}^n_Z,G$ for all $1 \leq i,l \leq n$ and $i \neq l$.
\end{assump}
By Assumption~\ref{assump:comm-dag}, we need to learn one DAG structure for all observations' distributions. This is reasonable in our example as different users tend to visit websites in the same order. For $G_X$ in Assumption~\ref{assump:comm-dag}, the DAG property ensures a class of orderings of $[d_X]$, denoted as $\Pi^{\star}$, such that for any $\pi^{\star} \in \Pi^{\star}$, $\pi^{\star}_j < \pi^{\star}_k$ only if $k \notin \textrm{an}(j)$. For $1 \leq d \leq d_X$, the incomplete ordering $\{\pi_1,\ldots,\pi_d\}$ is consistent with $\Pi^{\star}$ if there exists $\pi^{\star} \in \Pi^{\star}$ such that $\pi^{\star}_1=\pi_1,\ldots,\pi^{\star}_d=\pi_d$. In other words, $\{\pi_1,\ldots,\pi_d\}$ is consistent with $\Pi^{\star}$ if it can be completed to a consistent ordering.

To simplify the dependency structure in $\mathbb{P} (\mathbf{X}^{(i)} \mid \mathcal{D}^n_Z,G )$, we assume the following about the node embedding function.

\begin{assump}[Node Embedding]
\label{assump:node-embedding}
We assume that there exists a node embedding function $h^{\star}_G$, such that for all $i=1,\ldots,n$, we have $\mathbf{Z}^{(i)}_{\star} \coloneqq h^{\star}_G (\mathbf{Z}^{(i)} ; \mathcal{D}^n_Z) \subseteq \mathbb{R}^{d^0_Z}$, $1\leq d^0_Z \leq d_Z$, and
\begin{equation*}
\mathbb{P} \left( \mathbf{X}^{(i)} \mid \mathcal{D}^n_Z, G \right) = \mathbb{P} \left( \mathbf{X}^{(i)} \mid \mathbf{Z}^{(i)}_{\star} \right).
\end{equation*}
We simplify the notation $h^{\star}_G (\mathbf{Z}^{(i)} ; \mathcal{D}^n_Z)$ to $h^{\star}_G(\mathbf{Z}^{(i)})$.
\end{assump}

Assumption~\ref{assump:node-embedding} states that a node embedding function $h^{\star}_G (\cdot)$ exists, simplifying the conditional structure of $\mathbf{X}^{(i)}$ on $G$ and $\mathcal{D}^n_Z$ to depend only on $h^{\star}_G(\mathbf{Z}^{(i)})$, facilitating estimation. An alternative to simplify the dependency of $\mathbf{X}^{(i)}$ on $G$ and $\mathcal{D}^n_Z$ is to let
\begin{equation}
\label{eq:altern-approach}
\mathbb{P} \left( \mathbf{X}^{(i)} \mid \mathcal{D}^n_Z, G \right) = \mathbb{P} \left( \mathbf{X}^{(i)} \mid \mathbf{Z}^{(i)} \right),
\end{equation}
as adopted by~\cite{kolar2010sparse}. While this approach seems natural, our approach based on Assumption~\ref{assump:node-embedding} has three advantages. First, while \eqref{eq:altern-approach} assumes $(\mathbf{X}^{(i)}, \mathbf{Z}^{(i)})$ is independent of $(\mathbf{X}^{(j)}, \mathbf{Z}^{(j)})$ for $i \neq j$, our approach allows dependency, which is more practical. Second, the original feature vector $\mathbf{Z}^{(i)}$ can be high-dimensional, complicating the penalized kernel smoothing method in Section~\ref{sec:methodolgy}. Our approach relies on the embedded feature $h_G^{\star}(\mathbf{Z}^{(i)})$, which is typically smaller. Finally, $\mathbf{Z}^{(i)}$ can be sparsely observed, as some users may lack certain feature records. In such cases, the approach based on \eqref{eq:altern-approach} suffers from reduced sample size. Our approach leverages network information to infer missing features from neighbors, allowing use of the entire dataset for estimation.



Given Assumptions~\ref{assump:comm-dag} and~\ref{assump:node-embedding}, we further specify $f_j(\cdot)$ analytically.
\begin{assump}[Binomial SEM]
\label{assump:binomial-sem}
Given $G$, $\mathcal{D}^n_Z$, and $G_X$, $\mathbf{X}^{(i)}$ is generated according to the following structural equation model (SEM) for all $i=1,\ldots,n$. 
For all $j \in V_X$, we have
\begin{equation*}
X^{(i)}_j \mid \mathcal{D}^n_Z, \mathbf{X}^{(i)}_{ \textrm{pa}(j) }, G \sim \text{Binomial} \left( T, p_j ( \eta^{(i)}_j ) \right),
\end{equation*}
where $T$ is the number of trials in the Binomial distribution, 
$p_j(\eta) = 1 / (1 + \exp \left( - \eta \right) )$ with $\eta>0$ is the probability, 
\begin{equation}
\label{eq:GLM-parent}
\eta^{(i)}_j = w_{jj} \left( h^{\star}_G \left( \mathbf{Z}^{(i)} \right) \right) + \sum_{ l \in \textrm{pa}(j) } w_{lj} \left( h^{\star}_G \left( \mathbf{Z}^{(i)} \right) \right) X_l,
\end{equation}
and $w_{jj} (\cdot),w_{lj} (\cdot): \mathbb{R}^{d^0_Z} \mapsto \mathbb{R}$ are smooth functions mapping embedded features to DAG weights.
\end{assump}

Assumption~\ref{assump:binomial-sem} states that a node's value, given the embedding of its feature vector and parents, follows a Binomial distribution with a probability determined by these embeddings. The parents' influence on the node's mean is modeled via a generalized linear model. By~\eqref{eq:GLM-parent} and the smoothness of $w_{jl}(\cdot)$, if $h^{\star}_G(\mathbf{Z}^{(i)})$ is close to $h^{\star}_G(\mathbf{Z}^{(k)})$, then observations $i$ and $k$ will have similar distributions. Although Assumption~\ref{assump:binomial-sem} is specific to Binomial data, the concepts in Assumptions~\ref{assump:comm-dag} and~\ref{assump:node-embedding} can be generalized to other data types by adjusting Assumption~\ref{assump:binomial-sem}.

\section{Methodology}
\label{sec:methodolgy}

We introduce our DAG estimation algorithm in Algorithm~\ref{alg:ODS-order}.
The algorithm consists of four steps: 1) estimate the node embedding function $\hat{h}_G(\cdot)$; 2) given the estimated node embedding function, estimate each node's neighbors via penalized kernel smoothing; 3) estimate the DAG ordering using overdispersion scores and the estimated neighborhood; 4) recover the DAG by repeating Step~2. Note that our Step~3 extends the overdispersion score from~\cite{park2015learning,park2017learning} to account for data heterogeneity.

\begin{algorithm}[t!]
\caption{DAG estimation with observation features}
\label{alg:ODS-order}
\begin{algorithmic}[1]
\STATE {\bfseries Input:} $\{ \mathbf{X}^{(i)}, \mathbf{Z}^{(i)} \}^n_{i=1}$ and the relationship network between observations $G$.
\STATE {\bfseries Output:} Estimated DAG ordering $\hat{\pi}$ and DAG edge set $\widehat{E}_{X} \in V_{X}\times V_{X}$.
\STATE {\bfseries Step 1:} Encode $\mathbf{Z}^{(i)}$ into a low-dimensional feature $\hat{h}_{G}(\mathbf{Z}^{(i)})$ with the estimated node embedding function $\hat{h}_G(\cdot)$ for all $i=1,\ldots,n$.
\STATE {\bfseries Step 2:} Estimate the neighborhood set of node $j$ as $\widehat{\mathcal{N}}(j)$ for all $j \in V_X$.
\STATE {\bfseries Step 3:} Estimate the ordering using re-weighted generalized overdispersion scores:
\STATE Calculate the overdispersion scores $\hat{\mathcal{S}} (1, j)$, $j=1,\dots,d_X$ using~\eqref{eq:ods-score-S1j}.
\STATE Let $\hat{\pi}_1=\argmin_j \hat{\mathcal{S}} (1, j)$.
\FOR{$v=2\dots,d_X-1$}{
\FOR{$j \in [d_X] \backslash \{ \hat{\pi}_1,\dots,\hat{\pi}_{v-1} \}$}{
\STATE Calculate the overdispersion score $\hat{\mathcal{S}} (v, j)$ using~\eqref{eq:ods-score-Svj}.
}
\ENDFOR
\STATE Let $\hat{\pi}_v = \argmin_{j} \hat{\mathcal{S}} (v, j)$.
}
\ENDFOR
\STATE The last element of the ordering $\hat{\pi}_{d_X}=[d_X] \backslash \{ \hat{\pi}_1,\dots,\hat{\pi}_{d_X-1} \}$.
\STATE Let $\hat{\pi}=(\hat{\pi}_1,\dots,\hat{\pi}_{d_X})$.
\STATE {\bfseries Step 4:} Get the directed graph $\widehat{E}_{X}=\cup_{m=\{2,3,\cdots,d_X\}}\widehat{D}_{m}$, where $\widehat{D}_{m}$ is the estimated parent set of node $\hat{\pi}_m$.
\STATE {\bfseries Return:} $\hat{\pi}$ and $\widehat{E}_{X}$.
\end{algorithmic}
\end{algorithm}

\subsection{Estimation of Node Embedding Function}

By Assumption~\ref{assump:node-embedding}, we assume a node embedding function that simplifies the dependency structure of observations on features and reduces the feature dimension. Thus, our first step of Algorithm~\ref{alg:ODS-order} is to estimate this function. Given the relationship network $G$, our aim is to find a function $\hat{h}_{G}(\cdot)$ that encodes network information into the transformed feature vectors $\{\hat{h}_{G}(\mathbf{Z}^{(i)})\}^n_{i=1}$. Users $i,j$ close to the graph $G$ should have embedded features $\hat{h}_{G}(\mathbf{Z}^{(i)}),\hat{h}_{G}(\mathbf{Z}^{(j)})$ closer than raw feature vectors $\mathbf{Z}^{(i)},\mathbf{Z}^{(j)}$. Next, we explain how to estimate linear and nonlinear embeddings. To simplify the notation, we use $\hat{\bm{h}}^{(i)}$ to denote the embedding $\hat{h}_G \left(\mathbf{Z}^{(i)}\right)$.

\subsubsection{Linear Embedding Function Estimation}
\label{sec:linear}

Assuming $h^{\star}_{G}(\cdot)$ is linear, it can be represented by a projection matrix: $h_{G}^{\star}(\cdot) = \langle \bm{F}^{\star},\cdot \rangle$. We follow the method of \citet{zhao2022dimension} to estimate $\bm{F}^{\star}$ which has good theoretical properties under certain conditions. Given a graph $G = (V,E)$, let $W^0 \in \mathbb{R}^{n \times n}$ be its adjacency matrix with $w^0_{ij}=w^0_{ji}=1$ if $(i,j) \in E$ and zero otherwise. \citet{zhao2022dimension} seeks $\bm{F} \in \mathbb{R}^{d_{Z}\times d_Z^0}$ such that small $\Vert \bm{F}^{\top}\mathbf{Z}^{(i)} - \bm{F}^{\top}\mathbf{Z}^{(j)} \Vert$ corresponds to $w^0_{ij}=1$. To ensure identifiability, $\bm{F}$ is restricted to $\Omega_{\bm{A}} \coloneqq \{\bm{F}: \bm{F}^{\top} \bm{A} \bm{F} = \bm{I} \} \subseteq \mathbb{R}^{d_Z\times d_Z^0}$ where $\bm{A}\in \mathbb{R}^{d_Z \times d_Z}$ is a user-chosen positive definite matrix with bounded eigenvalues. 
The matrix $\bm{F}^{\star}$ is estimated as 
\begin{equation*}
\widehat{\bm{F}} = \arg \max_{\bm{F}\in\Omega_{\bm{A}}} \, \frac{1}{n(n-1)}\sum_{i\neq j}(1-w^0_{ij})\left\Vert \bm{F}^{\top}\left( \mathbf{Z}^{(i)} - \mathbf{Z}^{(j)}\right) \right\Vert^2,
\end{equation*}
which has an analytical solution as $\widehat{\bm{F}} = \bm{A}^{-1/2}\widehat{\bm{\Psi}}$, where $\widehat{\bm{\Psi}}$ is the matrix of eigenvectors associated with the $d_Z^0$ largest eigenvalues of $\bm{A}^{-1/2}\widehat{\bm{C}} \bm{A}$ and
\begin{equation*}
\widehat{\bm{C}}=\frac{1}{n(n-1)}\sum_{i\neq j}(1-w^0_{ij})\left(\mathbf{Z}^{(i)}-\mathbf{Z}^{(j)}\right)\left(\mathbf{Z}^{(i)}-\mathbf{Z}^{(j)}\right)^{\top}
\end{equation*}
If not specified, $\bm{A} = \textrm{Cov}(\mathbf{Z})$. The node embedding function is estimated as $\hat{h}_G(\cdot) = \langle \widehat{\bm{F}}, \cdot \rangle$.

\subsubsection{Nonlinear Embedding Function Estimation}
\label{sec:nonlinear}

If $h^{\star}_G(\cdot)$ is not required to be linear, we use Graph Auto-Encoders (GAEs)~\citep{kipf2016variational} to encode network information. GAEs are unsupervised frameworks that encode node features into a latent space via an encoder and reconstruct graph data via a decoder. The encoder uses graph convolutional layers to obtain low-dimensional node embeddings. The decoder computes pair-wise distances from these embeddings and reconstructs the adjacency matrix after a non-linear activation layer. The network is trained by minimizing the discrepancy between the real and reconstructed adjacency matrices. GAEs can output embedded features from node features or node embeddings encoding network topology without input features.

\subsection{Estimation of Neighborhood Set}
\label{sec:neighbor-selection}

The neighborhood set of node $j \in V_X$, denoted by $\mathcal{N}(j)$, is the minimal subset of $V_X$ such that 
\begin{equation*}
X^{(i)}_j \independent \mathbf{X}^{(i)}_{V_X \backslash \mathcal{N}(j)} \mid h^{\star}_G \left( \mathbf{Z}^{(i)} \right), \mathbf{X}^{(i)}_{\mathcal{N}(j)}
\qquad \text{ for all } i \in [n].
\end{equation*}
Thus, $\textrm{pa}(j) \subseteq \mathcal{N}(j)$. If we knew the set $\mathcal{N}(j)$, this would allow us to reduce the search space for $\textrm{pa}(j)$ to the nodes in $\mathcal{N}(j)$. Since estimating the set $\mathcal{N}(j)$ is easier than estimating $\textrm{pa}(j)$, due to ignoring the directional information, the set $\mathcal{N}(j)$ is first estimated in Step~2 of Algorithm~\ref{alg:ODS-order}.

Following~\cite{yang2012graphical} and \cite{park2017learning}, we recast the estimation of the neighborhood set as a variable selection problem. 
Under Assumption~\ref{assump:binomial-sem}, \citet{park2017learning} proposed to estimate $\mathcal{N}(j)$ in an i.i.d.~setting by estimating $\bm{\gamma}^{(i)}_j=(\bm{\gamma}^{(i)}_{1j},\ldots,\bm{\gamma}^{(i)}_{dj})^{\top}$, defined as the minimizer of 
\begin{equation}
\label{eq:expected-Xi}
\min_{\bm{\gamma} \in \Gamma_j:  } \mathbb{E}_{ \bm{X}^{(i)} } \left[ \left( T-X^{(i)}_j \right) \left( \gamma_j + \sum_{l \in \mathcal{N}(j) }  \gamma_{l}  X^{(i)}_l \right) + T \log \left( 1 + \exp \left(  - \gamma_j - \sum_{l \in \mathcal{N}(j) }  \gamma_{l}  X^{(i)}_l  \right)  \right) \right],
\end{equation}
where $\Gamma_j=\{ \bm{\gamma} \in \mathbb{R}^{d}  :  \gamma_l = 0 \text{ for all } l \notin  \mathcal{N}_j \cup \{ j \} \}$. The estimator of $\bm{\gamma}^{(i)}_j$ is obtained by minimizing the following penalized objective:
\begin{equation*}
\hat{\bm{\gamma}}^{(i)}_j = \arg \min_{\bm{\gamma} \in \mathbb{R}^{d_X}} l_j (\gamma) + \lambda \sum_{l \neq j} \vert \gamma_l \vert,
\end{equation*}
where
\begin{equation*}
l_j (\bm{\gamma}) = \frac{1}{n} \sum^n_{i=1} \left( T-X^{(i)}_j \right) \left( \gamma_j + \sum_{l \in \mathcal{N}(j) }  \gamma_{l}  X^{(i)}_l \right) + T \log \left( 1 + \exp \left(  - \gamma_j - \sum_{l \in \mathcal{N}(j) }  \gamma_{l}  X^{(i)}_l  \right)  \right).
\end{equation*}
The neighborhood set of node $j$ is then obtained as the support of $\hat{\bm{\gamma}}^{(i)}_j$. Under mild conditions, $\textrm{Supp}(\hat{\bm{\gamma}}^{(i)}_j)=\mathcal{N}(j)$ with high probability \citep{park2017learning}.

In contrast to the setting in~\cite{park2017learning}, the observations $\{\bm{X}^{(i)}\}_{i=1}^n$ in our set-up depend on the covariates $\{\mathbf{Z}^{(i)}\}_{i=1}^n$. As a result, the loss function $l_j(\cdot)$ does not approximate $\mathbb{E}_{X^{(i)}}[\cdot]$ well. To overcome this challenge, we use the key observation that when $h^{\star}_G \left( \mathbf{Z}^{(i)} \right) \approx h^{\star}_G \left( \mathbf{Z}^{(j)} \right)$, we would have $\bm{\gamma}^{(i)}_j \approx \bm{\gamma}^{(j)}_j$. This insight suggests employing a penalized kernel smoothing estimator to approximate the expected loss under $\mathbb{E}_{X^{(i)}}[\cdot]$ by assigning higher weights to samples with node embeddings similar to $h^{\star}_G \left( \mathbf{Z}^{(i)} \right)$, instead of distributing equal weights to all samples~\citep{kolar2010sparse}.

We first construct an estimate of the expected loss in~\eqref{eq:expected-Xi}. Let 
\begin{gather*}
\mathbf{b}^{(i)}_j=\left( b^{(i)}_{1j}, \dots, b^{(i)}_{d_X j} \right)^{\top} \in \mathbb{R}^{d_X}, \qquad
\mathbf{b}^{(i)}_{- j,j}= \left( b^{(i)}_{1j}, \ldots, b^{(i)}_{j-1, j} ,b^{(i)}_{j+1, j},\ldots,b^{(i)}_{d_X j} \right)^{\top},\\
\mathbf{X}^{(i)}_{-j}=\left( X^{(i)}_1,\ldots,X^{(i)}_{j-1},X^{(i)}_{j+1},\ldots,X^{(i)}_{d_X} \right)^{\top}.
\end{gather*}
For $j \in [d_X]$ and $i \in [n]$, a local estimate of~\eqref{eq:expected-Xi} is 
\begin{multline}
\label{eq:expected-Xi:local}
\mathcal{L}^{(i)}_j \left( \mathbf{b}^{(i)}_j \right) \coloneqq \sum_{k =1 }^n \frac{ K_{\tau_1} \left( \hat{\bm{h}}^{(i)} - \hat{\bm{h}}^{(k)} \right) }{ \sum^n_{l=1} K_{\tau_1} \left( \hat{\bm{h}}^{(i)} - \hat{\bm{h}}^{(l)} \right) } \times  \left\{ \left( T-X^{(k)}_j \right) \left(  b^{(i)}_{jj} + \left\langle \mathbf{b}^{(i)}_{-j,j} , \mathbf{X}^{(k)}_{-j} \right\rangle  \right) \right. \\
+ \left. T \log \left( 1 + \exp \left( -  b^{(i)}_{jj} - \left\langle \mathbf{b}^{(i)}_{-j,j} , \mathbf{X}^{(k)}_{-j} \right\rangle \right)  \right) \right\}
\end{multline}
where $K_{\tau}(\mathbf{x})=K(\Vert \mathbf{x} \Vert/\tau)$, $K(\cdot)$ is a symmetric positive real valued function that defines local weights, and $\tau_1>0$ is the bandwidth. Throughout, we use $K(u)=\exp(-|u|)$. Minimizing the local loss in \eqref{eq:expected-Xi:local} gives us an estimate of $\bm{\gamma}^{(i)}_j$. In \eqref{eq:expected-Xi:local}, we assign a weight to observation $k$ based on the similarity between $h^{\star}_G \left( \mathbf{Z}^{(i)} \right)$ and $h^{\star}_G \left( \mathbf{Z}^{(k)} \right)$. Based on \eqref{eq:expected-Xi:local}, we estimate $\bm{\gamma}^{(i)}_{lj}$, $l,j \in [d_X]$ and $i \in [n]$, by minimizing the following penalized objective:
\begin{equation}
\label{eq:loss-nb-moral}
\begin{aligned}
\widehat{\mathbf{B}}_j \coloneqq \argmin_{\mathbf{B}_j \in \mathbb{R}^{d_X \times n } }
\mathcal{L}_j \left(  \mathbf{B}_j \, ; \mathcal{D}^n \right) \coloneqq & \sum_{ i =1 }^n \mathcal{L}^{(i)}_j \left( \mathbf{b}^{(i)}_j \right) + \lambda^j  \sum_{l \neq j}\left\Vert \mathbf{b}_{lj} \right\Vert_2,
\end{aligned}
\end{equation}
where $\mathbf{B}_j = \left[ \mathbf{b}^{(1)}_j,\ldots,\mathbf{b}^{(n)}_j \right] \in \mathbb{R}^{d_X \times n}$ and $\mathbf{b}_{lj}=\left( b^{(1)}_{lj},\ldots,b^{(n)}_{lj} \right)^{\top}$ is the $l$-th row of $\mathbf{B}_j$. The second term in~\eqref{eq:loss-nb-moral} is the penalty that encourages group-structured sparsity~\citep{yuan2006model}. Since for $l \notin \mathcal{N}(j) \cup \{ j \}$, $\{ \bm{\gamma}^{(i)}_{lj} \}^n_{i=1}$ are all zero, the group lasso penalty encourages row sparsity in $\widehat{\mathbf{B}}_j$. The penalty parameter $\lambda^j > 0$ controls the sparsity level of the neighborhood set of node $j$.
Finally, we estimate the neighborhood of node $j$ by
$
\widehat{\mathcal{N}} (j) \coloneqq \left\{ l \, : \, \left\Vert \hat{\mathbf{b}}_{l j} \right\Vert_2 > 0 \right\}
$.

\subsubsection{Approximate Optimization Algorithm for Solving~\eqref{eq:loss-nb-moral}}
\label{sec:approx-opt}


Although it is feasible to solve directly~\eqref{eq:loss-nb-moral}, in practice, a large number of observations $n$ can present significant challenges for both storage and computation. We propose an approximate solution to the original problem~\eqref{eq:loss-nb-moral} that is suitable for large-scale datasets. Our approximate solution is closely related to the concept of binning in the nonparametric statistics literature~\citep{scott1981using, silverman1982algorithm, fan1994fast, wand1994fast}. However, while binning is effective for univariate or low-dimensional cases, our node embedding may have relatively high dimensions. To address this issue, we extend the binning concept to clustering. Instead of dividing the space into bins, we cluster samples into groups and fit one parameter for each group rather than for each sample. By doing so, we reduce the computational complexity from being proportional to the sample size to being proportional to the group size. For the clustering method, we use $K$-means clustering in this paper~\citep{hartigan1979algorithm}, but other clustering methods can also be applied here, such as spectral clustering~\citep{ng2001spectral}.

Given embeddings $\{\hat{\bm{h}}^{(i)}\}^n_{i=1}$, we apply $k$-means clustering to divide the observations into $M$ groups $\{\mathcal{A}_m\}^M_{m=1}$, with $\cup^M_{m=1}\mathcal{A}_m=[n]$ and centers $\{\mathbf{c}_m\}^M_{m=1}$. We replace $\hat{\bm{h}}^{(i)}$ with $\mathbf{c}_m$ for $i \in \mathcal{A}_m$, reducing the parameters needed to $M$ instead of $n$. When $M=n$, we recover the original problem. 
We let
\begin{equation*}
\alpha^{(m)}_i = \frac{ K_{\tau_1} \left( \Vert \hat{\bm{h}}^{(i)} - \mathbf{c}_m \Vert \right)   }{ \sum^n_{l=1} K_{\tau_1} \left( \Vert \hat{\bm{h}}^{(l)} - \mathbf{c}_m \Vert \right) }\quad \text{for all } 1 \leq i \leq n, \, 1 \leq m \leq M.
\end{equation*}
Besides, for a given $j \in [d]$,
let $\mathbf{b}^{(m)}_j = (b^{(m)}_{1j},\ldots,b^{(m)}_{d_X,j})^{\top} \in \mathbb{R}^{d_X}$, and $\mathbf{B}_j=[\mathbf{b}^{(1)}_j,\ldots,\mathbf{b}^{(M)}_j  ] \in \mathbb{R}^{d_X \times M}$. We use $\mathbf{b}_{lj}=\left( b^{(1)}_{lj},\ldots,b^{(M)}_{lj} \right)^{\top}$ to denote the $l$-th row of matrix $\mathbf{B}_j$.
We solve the following optimization problem as an approximation to~\eqref{eq:loss-nb-moral}:
\begin{equation}
\label{eq:cond-bin-glm-obj}
\min_{ \mathbf{B} } F (\mathbf{B}) \coloneqq  l( \mathbf{B}) + \lambda \sum^{d_X}_{j=2} \Vert \mathbf{B}_{j \cdot} \Vert_2,
\end{equation}
where
\begin{gather*}
l( \mathbf{B} ) \coloneqq \frac{1}{M} \sum^M_{m=1} \sum^n_{i=1} \alpha^{(m)}_i \left\{ \left( T - X^{(i)}_j \right) \left( b^{(m)}_{jj} + \left\langle \mathbf{b}^{(m)}_{-j,j}, \mathbf{X}^{(i)}_{-j}  \right\rangle \right)  \right. \\
\left. + T \log \left( 1 + \exp \left( - b^{(m)}_{jj} - \left\langle \mathbf{b}^{(m)}_{-j,j} , \mathbf{X}^{(i)}_{-j} \right\rangle \right) \right) \right\}.
\end{gather*}
Thus, the number of parameters to estimate is $Md_X$, which is significantly smaller than $n d_X$ when $n$ is large. Observe that~\eqref{eq:cond-bin-glm-obj} features a composite structure of smooth convex loss combined with a non-smooth convex penalty, we can then use a proximal-gradient method to solve~\eqref{eq:cond-bin-glm-obj}~\citep{parikh2014proximal}.

\subsection{Determine the Order}
\label{sec:dt-order}

In Step~3 of Algorithm~\ref{alg:ODS-order}, we determine the ordering of the DAG by modifying the method of~\cite{park2017learning} to fit our context. The concept involves creating a series of overdispersion scores that depend on the conditional mean and conditional variance, and then determining the order by sequentially selecting the node with the lowest overdispersion score. Unlike~\cite{park2017learning}, in this paper, the conditional mean and variance differ across samples, adding complexity to the estimation. To address this, we employ a similar kernel smoothing technique as described in Section~\ref{sec:neighbor-selection}.

We start by defining the overdispersion scores.
By Assumption~\ref{assump:binomial-sem}, we have the conditional expectation of a node $j$ as $\mathbb{E} \left[ X^{(i)}_j \mid h^{\star}_G\left(\mathbf{Z}^{(i)}\right),\mathbf{X}^{(i)}_{\textrm{pa}(j)}  \right] = T p_j ( \eta^{(i)}_j )$ and the conditional variance as $\mathbb{V} \left[ X^{(i)}_j \mid h^{\star}_G\left(\mathbf{Z}^{(i)}\right),\mathbf{X}^{(i)}_{\textrm{pa}(j)}  \right] = T p_j ( \eta^{(i)}_j ) \left( 1 - p_j ( \eta^{(i)}_j ) \right)$. In particular, there is a quadratic relation between conditional mean and variance as
\begin{equation*}
\mathbb{V} \left[ X^{(i)}_j \mid h^{\star}_G\left(\mathbf{Z}^{(i)}\right),\mathbf{X}^{(i)}_{\textrm{pa}(j)}  \right] = \mathbb{E} \left[ X^{(i)}_j \mid h^{\star}_G\left(\mathbf{Z}^{(i)}\right),\mathbf{X}^{(i)}_{\textrm{pa}(j)}  \right] - \frac{1}{T} \left(  \mathbb{E} \left[ X^{(i)}_j \mid h^{\star}_G\left(\mathbf{Z}^{(i)}\right),\mathbf{X}^{(i)}_{\textrm{pa}(j)}  \right]  \right)^2.
\end{equation*}
By defining $\omega^i_j \coloneqq \omega_j \left( h^{\star}_G\left(\mathbf{Z}^{(i)}\right), \mathbf{X}^{(i)}_{\textrm{pa}(j)} \right) \coloneqq \left( 1 - \frac{1}{T} \mathbb{E} \left[ X^{(i)}_j \mid h^{\star}_G\left(\mathbf{Z}^{(i)}\right), \mathbf{X}^{(i)}_{\textrm{pa}(j)}  \right]   \right)^{-1}$, we have 
\begin{equation*}
\mathbb{V} \left[ \omega^i_j X^{(i)}_j \mid h^{\star}_G\left(\mathbf{Z}^{(i)}\right),\mathbf{X}^{(i)}_{\textrm{pa}(j)}  \right] = \mathbb{E} \left[ \omega^i_j X^{(i)}_j \mid h^{\star}_G\left(\mathbf{Z}^{(i)}\right),\mathbf{X}^{(i)}_{\textrm{pa}(j)}  \right].
\end{equation*}
Based on this, we define the following overdispersion scores that can help identify an ordering that is consistent with $\Pi^{\star}$. For $j \in [d_X]$, we define
\begin{align*}
\omega_{1j}\left(\mathbf{Z}^{(i)}\right) & \coloneqq \left( 1 - \frac{1}{T} \mathbb{E} \left[ X^{(i)}_j \mid \mathbf{Z}^{(i)} \right] \right)^{-1}, \\
\mathcal{S}^i (1,j) & \coloneqq \omega_{1j}^2\left(\mathbf{Z}^{(i)}\right) \mathbb{V} \left[ X^{(i)}_j \mid \mathbf{Z}^{(i)} \right] - \omega_{1j}\left(\mathbf{Z}^{(i)}\right) \mathbb{E} \left[ X^{(i)}_j \mid \mathbf{Z}^{(i)} \right], \\
\mathcal{S} (1,j) & \coloneqq \frac{1}{n} \sum^n_{i=1} \mathcal{S}^i (1,j).
\end{align*}
Following the proof of Theorem 5 in~\cite{park2017learning}, it can be demonstrated that if we set $j^{\star} = \argmin_j \mathcal{S} (1,j)$, then $\textrm{pa}(j^{\star}) = \emptyset$. Thus, the root node is identified by checking $\mathcal{S} (1,j)$.
Furthermore, for $v \geq 2$ and an incomplete ordering $\pi_{1:(v-1)}=\{\pi_1,\ldots,\pi_{v-1}\}$ that is consistent with $\Pi^{\star}$, we define
\begin{align*}
\omega_{vj}\left(\mathbf{Z}^{(i)}\right) & \coloneqq \left( 1 - \frac{1}{T} \mathbb{E} \left[ X^{(i)}_j \mid \mathbf{Z}^{(i)}, \mathbf{X}^{(i)}_{\pi_{1:(v-1)}} \right] \right)^{-1}, \\
\mathcal{S}^i (v,j) & \coloneqq \omega_{vj}^2\left(\mathbf{Z}^{(i)}\right) \mathbb{V} \left[ X^{(i)}_j \mid \mathbf{Z}^{(i)},\mathbf{X}^{(i)}_{\pi_{1:(v-1)}} \right] - \omega_{vj}\left(\mathbf{Z}^{(i)}\right) \mathbb{E} \left[ X^{(i)}_j \mid \mathbf{Z}^{(i)},\mathbf{X}^{(i)}_{\pi_{1:(v-1)}} \right], \\
\mathcal{S} (v,j) & \coloneqq \frac{1}{n} \sum^n_{i=1} \mathcal{S}^i (v,j),
\end{align*}
for $j \in [d_X] \backslash \{\pi_1,\ldots,\pi_{v-1}\}$.
Letting $\pi_v=\argmin_j \mathcal{S} (v,j)$ ensures $\pi_{1:v}=\{\pi_1,\ldots,\pi_v,\}$ is consistent with $\Pi^{\star}$. Thus, using these conclusions and induction, we can sequentially estimate overdispersion scores and the ordering.

We then discuss how to estimate overdispersion scores.
We estimate overdispersion scores by first constructing conditional mean and variance estimates:
\begin{gather*}
\hat{\mathbb{E}} \left[ X^{(i)}_j \mid \mathbf{Z}^{(i)} \right] =  \sum^n_{l=1} X^{(l)}_j \theta^i_l, \quad\quad
\hat{\mathbb{V}} \left[ X^{(i)}_j \mid \mathbf{Z}^{(i)}  \right] =  \sum^n_{l=1}  \left( X^{(i)}_l  - \hat{\mathbb{E}} \left[ X^{(i)}_j \mid \mathbf{Z}^{(i)} \right] \right)^2 \theta^i_l , \\
\quad \text{ where } \theta^i_l = \frac{ K_{\tau_2} \left( \hat{\bm{h}}^{(i)} - \hat{\bm{h}}^{(l)} \right)  }{ \sum^n_{k=1} K_{\tau_2} \left( \hat{\bm{h}}^{(i)} - \hat{\bm{h}}^{(k)} \right) },
\end{gather*}
and $K_{\tau}(\cdot)$ is defined in Section~\ref{sec:neighbor-selection} with $\tau_2>0$ being the bandwidth. 
We utilize the similar idea of kernel smoothing as in Section~\ref{sec:neighbor-selection} to borrow information from neighboring samples when estimating the conditional mean and variance of a sample.
We then define $\hat{\omega}_{1j} \left( \mathbf{Z}^{(i)} \right) = \left( 1- \frac{1}{T} \hat{\mathbb{E}} \left[ X^{(i)}_j \mid \mathbf{Z}^{(i)} \right]   \right)^{-1}$ and
\begin{equation}
\label{eq:ods-score-S1j}
\hat{\mathcal{S}} (1, j) \coloneqq \frac{1}{n} \sum_{i=1}^{n} \left\{ \hat{\omega}^2_{1j} \left( \mathbf{Z}^{(i)} \right) \hat{\mathbb{V}} \left[ X^{(i)}_j \mid \mathbf{Z}^{(i)} \right] - \hat{\omega}_{1j} (\mathbf{Z}^{(i)}) \hat{\mathbb{E}} \left[ X^{(i)}_j \mid \mathbf{Z}^{(i)}  \right] \right\}.
\end{equation}
The estimate $\hat{\mathcal{S}} (1, j)$ is used to the root node as $\hat{\pi}_1 = \argmin_j \hat{\mathcal{S}} (1, j)$.

Suppose that we have obtained the set $\{ \hat{\pi}_1,\dots,\hat{\pi}_{v-1} \}$ for $v \geq 2$. For any $j \in [d_X] \setminus \{ \hat{\pi}_1,\dots,\hat{\pi}_{v-1} \}$, we define the candidate parent set of node $j$ as $\widehat{C}_{vj} = \widehat{\mathcal{N}}(j) \cap \{ \hat{\pi}_1,\dots,\hat{\pi}_{v-1} \}$.
For any $j \in V_X$ and $S \subseteq [d_X] \backslash \{j\}$, the estimate of the conditional expectation and variance given the set $S$ is
\begin{gather*}
\hat{\mu}^i_{j, \mathbf{x}_S} =
\hat{\mathbb{E}} \left[ X^{(i)}_j \mid \mathbf{Z}^{(i)},\mathbf{X}^{(i)}_S=\mathbf{x}_S \right] = \sum^n_{l=1} X^{(i)}_j \cdot \theta^i_{l,\mathbf{x}_S} \cdot \mathds{1} \{ \mathbf{X}^{(l)}_S = \mathbf{x}_S \}, \\
\hat{\mathbb{V}} \left[ X^{(i)}_j \mid \mathbf{Z}^{(i)},\mathbf{X}^{(i)}_S=\mathbf{x}_S  \right] = \sum^n_{l=1} \left(  X^{(i)}_l  - \hat{\mu}^i_{j, \mathbf{x}_S} \right)^2 \cdot \theta^i_{l,\mathbf{x}_S} \cdot \mathds{1} \{ \mathbf{X}^{(l)}_S = \mathbf{x}_S \},
\end{gather*}
where $\mathbf{X}^{(i)}_S$ is a subvector of $\mathbf{X}^{(i)}$ and 
\begin{equation*}
\theta^i_{l,\mathbf{x}_S} = \frac{ K_{\tau_2} \left( \hat{\bm{h}}^{(i)} - \hat{\bm{h}}^{(l)} \right) \cdot \mathds{1} \{ \mathbf{X}^{(l)}_S = \mathbf{x}_S \}  }{ \sum^n_{k=1} K_{\tau_2} \left( \hat{\bm{h}}^{(i)} - \hat{\bm{h}}^{(k)} \right) \cdot \mathds{1} \{ \mathbf{X}^{(k)}_S = \mathbf{x}_S \} }.
\end{equation*}
Let $\mathcal{X}(S)=\{0,1,\ldots,T\}^{\vert S \vert}$. For $\mathbf{x}_S \in \mathcal{X}(S)$, let $n(\mathbf{x}_S) \coloneqq \sum^n_{i=1} \mathds{1}(\mathbf{X}^{(i)}_S=\mathbf{x}_S)$ denote the conditional sample size and $n_S=\sum_{ \mathbf{x}_S \in \mathcal{X}(S)} n(\mathbf{x}_S) \mathds{1} ( n(\mathbf{x}_S) \geq n_0 )$ denote the truncated conditional sample size, where $n_0$ is a tuning parameter with $1 \leq n_0 \leq n$. With this notation, for $v=2,\dots,d_X-1$ and the candidate parent set $\hat{C}_{vj}$, we define $\hat{\omega}_{vj} \left( \mathbf{Z}^{(i)}, \mathbf{x} \right) = \left( 1- \frac{1}{T} \hat{\mathbb{E}} \left[ X^{(i)}_j \mid \mathbf{Z}^{(i)},\mathbf{X}^{(i)}_{\hat{C}_{vj}}=\mathbf{x} \right]   \right)^{-1}$ and 
\begin{multline}\label{eq:ods-score-Svj}
\hat{\mathcal{S}} (v, j) \coloneqq \frac{1}{n} \sum_{i=1}^{n} \sum_{\mathbf{x} \in \mathcal{X}_{\hat{C}_{vj}} } \frac{n(\mathbf{x})}{n_{ \hat{C}_{vj} }} \left\{ \hat{\omega}^2_{vj} (\mathbf{Z}^{(i)}, \mathbf{x}) \cdot \hat{\mathbb{V}} \left[ X^{(i)}_j \mid \mathbf{Z}^{(i)},\mathbf{X}^{(i)}_{\hat{C}_{vj}}=\mathbf{x}  \right] - \right. \\ 
\left. \hat{\omega}_{vj} (\mathbf{Z}^{(i)}, \mathbf{x}) \cdot \hat{\mathbb{E}} \left[ X^{(i)}_j \mid \mathbf{Z}^{(i)},\mathbf{X}^{(i)}_{\hat{C}_{vj}}=\mathbf{x}  \right] \right\},
\end{multline}
where the summation over the set $\mathcal{X}_{ \hat{C}_{vj} } \coloneqq \left\{ \mathbf{x}_{ \hat{C}_{vj} } \in \mathcal{X} ( \hat{C}_{vj} ): n(\mathbf{x}_{ \hat{C}_{vj} }) \geq n_0 \right\}$ is to ensure that we have enough samples to estimate element of the overdispersion score.
In this paper, we choose $n_0=2$. Ultimately, we determine the subsequent element of the ordering estimate as $\hat{\pi}_v=\argmin_j \hat{\mathcal{S}} (v, j)$.

\subsection{Recover the DAG}
\label{sec:directed-graph}

With estimated ordering $\hat{\pi}$, we can reconstruct the DAG using a penalized estimation procedure, similar to neighborhood selection in Section~\ref{sec:neighbor-selection} and originally proposed in~\cite{shojaie2010penalized}. To recover the DAG, we estimate the parent set of each node. Given $\pi^{\star} \in \Pi^{\star}$ and $v \geq 2$, to find the parent set of node $\pi^{\star}_v$, we search among $\{\pi^{\star}_1, \ldots, \pi^{\star}_{v-1}\}$. We use $\hat{\pi}$ instead of $\pi^{\star}$ for this search.

For $v=2,\dots,d_X$, we define $\hat{\mathbf{A}}_v \in \mathbb{R}^{v \times n}$ as a minimizer of  the following objective:
\begin{multline}
\label{eq:loss-directed}
\min_{\mathbf{A}_v \in \mathbb{R}^{v \times n } }
\sum_{i^{\prime} =1 }^{n}   \sum_{ i =1 }^{n} 
\frac{ K_{\tau_1} \left( \hat{\bm{h}}^{(i)} - \hat{\bm{h}}^{(i^{\prime})} \right) }{ \sum^n_{l=1} K_{\tau_1} \left( \hat{\bm{h}}^{(i)} - \hat{\bm{h}}^{(l)} \right) } \times
 \left\{ \left( T-X^{(i)}_{\hat{\pi}_v} \right) \left(  a^{(i^{\prime})}_{vv} + \sum^{v-1}_{u=1} a^{(i^{\prime})}_{u v} X^{(i)}_{\hat{\pi}_u} \right) \right. \\
+ \left. T \log \left( 1 + \exp \left( -  a^{(i^{\prime})}_{vv} - \sum^{v-1}_{u=1} a^{(i^{\prime})}_{u v} X^{(i)}_{\hat{\pi}_u}  \right)  \right) \right\} + \lambda^{\hat{\pi}_v} \sum^{v-1}_{u=1} \left\Vert \mathbf{a}_{uv} \right\Vert_2,
\end{multline}
where $\mathbf{A}_v = \left[ \mathbf{a}^{(i)}_v \right]_{i=1}^{n} \in \mathbb{R}^{v \times n}$, $\mathbf{a}^{(i)}_v=( a^{(i)}_{1v}, \dots, a^{(i)}_{vv} )^{\top} \in \mathbb{R}^{v}$, and $\mathbf{a}_{uv}$ is the $u$-th row of $\mathbf{A}_v$. Finally, based on $\hat{\mathbf{A}}_v$ we estimate the parent set of node $\hat{\pi}_v$ as
\begin{equation*}
\widehat{D}_v =
\widehat{\textrm{pa}} (\hat{\pi}_v) \coloneqq \left\{ 1 \leq u \leq v-1 \, : \, \left\Vert \hat{\mathbf{a}}_{u v} \right\Vert_2 > 0 \right\}.
\end{equation*}
The DAG estimate is $\widehat{E}_{X}=\cup_{m=2}^{d_X}\widehat{D}_{m}$. To solve~\eqref{eq:loss-directed}, we use the approximation technique from Section~\ref{sec:approx-opt}.

\section{Hyperparameter Tuning}
\label{sec:hyperparameter}

In this section, we present our approach for selecting the hyperparameters in our algorithm. There are four sets of hyperparameters that need to be determined: the bandwidth $\tau_1$ and the group lasso penalty parameters $\{\lambda_j\}^{d_X}_{j=1}$ in Step 2 and Step 4 of Algorithm~\ref{alg:ODS-order}, the bandwidth $\tau_2$ and the threshold sample size $n_0$ in Step 3 of Algorithm~\ref{alg:ODS-order}. In our experiments, we set $n_0=2$, and $\tau_1=\tau_2=n^{-1/5}$ as recommended in~\cite{kolar2010sparse}. In the following, we elaborate on the selection of $\{\lambda_j\}^{d_X}_{j=1}$.

\begin{algorithm}[t!]
\caption{Penalty Parameter Tuning for Heterogeneous Conditional Binomial DAG Learning}
\label{alg:lambda_tuning}
\begin{algorithmic}[1]
\STATE {\bfseries Input:} $\{\mathbf{X}^{(i)},\mathbf{Z}^{(i)}\}_{i=1}^{n} $.
\STATE {\bfseries Output:} Optimal parameter set $\{ \lambda^{\star}_{j}\}_{j=1}^{d_X}$ for loss~\eqref{eq:loss-nb-moral} or loss~\eqref{eq:loss-directed}.
\STATE Randomly partition $\{\mathbf{X}^{(i)},\mathbf{Z}^{(i)}\}_{i =1}^{n}$ into $q$ subsets with the same size.
\FOR{$j=1,2,\cdots,d_X$}{
\STATE Calculate the candidate grid of $\lambda^j$, denoted by $\lambda^j_{\textrm{seq}}$, by targeting $\lambda^j_{\textrm{min}}$ and $\lambda^j_{\textrm{max}}$ with~\eqref{eq:lambda_max}.
\FOR{$\lambda \in \lambda^j_{\textrm{seq}}$}{
\FOR{$l=1,2,\cdots,q$}{
\STATE Choose the $l$-th partition of $\{\mathbf{X}^{(i)},\mathbf{Z}^{(i)}\}_{i = 1}^{n}$, $\mathcal{I}_l$, as the validation set, and the remaining dataset, $\mathcal{I}_{-l}$, as the training set.
\STATE Get $\hat{\mathbf{B}}_j$ or $\hat{\mathbf{A}}_j$ by optimizing loss~\eqref{eq:loss-nb-moral} or loss~\eqref{eq:loss-directed}, respectively, with $\lambda$ on the training set.
\STATE Calculate the mean squared error(MSE) $\varepsilon_j(\lambda,l)$ by~\eqref{eq:mse_penalty}.
}
\ENDFOR
\STATE Denote $\varepsilon_j(\lambda,\cdot) = (\varepsilon_j(\lambda,1),\varepsilon_j(\lambda,2),\cdots,\varepsilon_j(\lambda,q))$, calculate the standard error of MSE as $\textrm{se}(\lambda)=\sqrt{\hat{\mathbb{V}}(\varepsilon_j(\lambda,\cdot))/q}$, where $\hat{\mathbb{V}}(\cdot)$ is the empirical variance.
}
\ENDFOR
\STATE Let $\varepsilon_{j,\min}=\min_{\lambda \in \lambda^j_{\textrm{seq}}} (1/q)\sum^q_{l=1}\varepsilon_j(\lambda,l)$ and $\lambda_{j,\min}=\argmin_{\lambda \in \lambda^j_{\textrm{seq}}} (1/q)\sum^q_{l=1}\varepsilon_j(\lambda,l)$, then we have
\begin{equation*}
\lambda^{\star}_{j} = \max \left\{ \lambda \in \lambda^j_{\textrm{seq}}: \frac{1}{q}\sum^q_{l=1}\varepsilon_j(\lambda,l) \leq \varepsilon_{j,\min} + \textrm{se}(\lambda_{j,\min})  \right\}.
\end{equation*}

}
\ENDFOR
\STATE {\bfseries Return:} $\{ \lambda^{\star}_{j}\}_{j=1}^{d_X}$.
\end{algorithmic}
\end{algorithm}
 
In Section~\ref{sec:neighbor-selection} and Section~\ref{sec:directed-graph}, the results of optimizing the $l_2$-penalized loss depend on the selection of penalty parameters $\{\lambda_j\}^{d_X}_{j=1}$, which determine the sparsity level of the estimated graph. Given the training data configuration $\mathcal{D}^n=\{ \mathbf{ X}^{(i)}, \mathbf{Z}^{(i)} \}^n_{i=1}$, we divide the dataset into $q$ equally sized subsets $\{\mathcal{I}_l\}^q_{l=1} \subseteq [n]$, where $\cup^q_{l=1} \mathcal{I}_l=[n]$. We define $\mathcal{I}_{-l} \coloneqq \cup^q_{k=1,k\neq l} \mathcal{I}_k$.
For a chosen node $j$, inspired by the grid setup of~\cite{meier2008group}, we define the grid of penalty parameters $\lambda^j_{\textrm{seq}}=\{\lambda^j_{\textrm{min}},\ldots,\lambda^j_{\textrm{max}}\}$, where $0 \leq \lambda^j_{\textrm{min}}<\ldots < \lambda^j_{\textrm{max}}$, by first letting
\begin{gather*}
e_{j,l} =\sum_{i=1}^{n} \frac{ K_{\tau_1} \left( \hat{\bm{h}}^{(i)} - \hat{\bm{h}}^{(l)} \right) }{ \sum^n_{k=1} K_{\tau_1} \left( \hat{\bm{h}}^{(k)} - \hat{\bm{h}}^{(l)} \right) } \times  X^{(i)}_l\left(X_j^{(i)} - \frac{1}{n}\sum_{i^{\prime}=1}^{n}X^{(i^{\prime})}_j\right), \quad \text{and} \\
\lambda^j_{\textrm{max}}=\frac{1}{n - 1}\max_{j\in[d_X]}\left\| (e_{j,1},e_{j,2},\cdots,e_{j,d_X}) \right\|_2. \numberthis  \label{eq:lambda_max}
\end{gather*}
We set $\lambda^j_{\textrm{min}}=\lambda^j_{\textrm{max}}/1000$ and distribute $\lambda^j_{\textrm{seq}}$ evenly between them. The gap distance, which is user-defined, controls the refinement of $\lambda^j_{\textrm{seq}}$.
We use mean squared error (MSE) for model evaluation. The model prediction is defined as
\begin{equation*}
\widehat{X}_j^{(i)}=Tp_j(\hat{\eta}^{(i)}_j)=\frac{T}{1+\exp{\left(-\hat{\eta}^{(i)}_j \right)}},
\end{equation*}
where $\hat{\eta}^{(i)}_j=\hat{b}^{(i)}_{jj}+\sum_{l\neq j}\hat{b}^{(i)}_{lj}X^{(i)}_l$ for loss~\eqref{eq:loss-nb-moral} and $\hat{\eta}^{(i)}_j = \hat{a}^{(i)}_{jj} + \sum_{l=1}^{j-1}\hat{a}^{(i)}_{lj}X^{(i)}_{\hat{\pi}_{l}}$ for loss~\eqref{eq:loss-directed}.
Thus, the MSE for node $j$ with penalty parameter $\lambda$ and partition $l$ is
\begin{equation}
\label{eq:mse_penalty}
\varepsilon_j(\lambda,l) \coloneqq \frac{1}{\vert \mathcal{I}_l \vert}\sum_{i \in \mathcal{I}_l } \left(X^{(i)}_{j} - \widehat{X}^{(i)}_{j} \right)^2.
\end{equation}
After getting $\varepsilon_j(\lambda,\cdot) = (\varepsilon_j(\lambda,1),\varepsilon_j(\lambda,2),\cdots,\varepsilon_j(\lambda,q))$, we calculate the standard error of MSE as $\textrm{se}(\lambda)=\sqrt{\hat{\mathbb{V}}(\varepsilon_j(\lambda,\cdot))/q}$, where $\hat{\mathbb{V}}(\cdot)$ is the empirical variance.
Then we choose $\lambda^{\star}_{j}$ as the largest value of $\lambda^j_{\textrm{seq}}$ such that the error is within 1 standard error of the minimum MSE that can be achieved. More specifically, let $\varepsilon_{j,\min}=\min_{\lambda \in \lambda^j_{\textrm{seq}}} (1/q)\sum^q_{l=1}\varepsilon_j(\lambda,l)$ and $\lambda_{j,\min}=\argmin_{\lambda \in \lambda^j_{\textrm{seq}}} (1/q)\sum^q_{l=1}\varepsilon_j(\lambda,l)$.
Then we have
\begin{equation*}
\lambda^{\star}_{j} = \max \left\{ \lambda \in \lambda^j_{\textrm{seq}}: \frac{1}{q}\sum^q_{l=1}\varepsilon_j(\lambda,l) \leq \varepsilon_{j,\min} + \textrm{se}(\lambda_{j,\min})  \right\}.
\end{equation*}
This rule selects $\lambda > \lambda_{j,\min}$ to obtain a sparser graph, which is generally preferred.

\section{Simulation Experiment}
\label{sec:simul-exp}

We validate our algorithm with simulations, demonstrating its superiority over previous QVF DAG models with covariates. To mimic real-world data, we generate the relationship network as in~\cite{weng2016community}. Experiments are divided based on whether the ground-truth embedding function is linear.

\subsection{Linear Setup}
\label{sec:exp-linear-setup}

We first generate the relationship network $G$ by the following procedure:
\begin{enumerate}

\item Let $L_{i}\in \{1,2\}$ be the user community label, generated from a Bernoulli distribution with $\mathbb{P}(L_{1})=\mathbb{P}(L_{2})=0.5$.

\item We generate $d_Z$-dimensional covariates $\mathbf{Z}^{(i)}\sim \mathcal{N}(\mu_{L_{i}},\Sigma)$ for $i=1,\ldots,n$, where $\bm{\Sigma}=(\sigma_{t_{1}t_{2}})$ with $\sigma_{t_{1}t_{2}}=0.4^{|t_{1}-t_{2}|}\mathds{1}(|t_{1}-t_{2}|<5)$ for $t_1,t_2\in [d_Z]$.

\item Given $(L_{i},L_{j},\mathbf{Z}_{ij})$ where $\mathbf{Z}_{ij} = \mathbf{Z}^{(i)} - \mathbf{Z}^{(j)}$, we generate $w^0_{ji}=w^0_{ij}\in\{0,1\}$ with
\begin{equation}
\label{eq:net_gen}
\mathbb{P}(w^0_{ij} \mid L_{i},L_{j},\mathbf{Z}_{ij})=\mathbb{P} \left( w^0_{ij} \mid L_{i}, L_{j} \right) \cdot \frac{\exp{(1 - \langle c_{\textrm{coef}},| \bm{F}^{\top} \mathbf{Z}_{ij}|_+\rangle})}{1+\exp{(1-\langle c_{\textrm{coef}},| \bm{F}^{\top} \mathbf{Z}_{ij}|_+\rangle})},
\end{equation}
where $|\cdot|_+$ denotes element-wise absolute value, and $\mathbb{P}(w^0_{ij} \mid L_{i},L_{j})$ is defined as $\mathbb{P}(w^0_{ij}=1 \mid L_{i}=L_{j})=a$ and $\mathbb{P}(w^0_{ij}=1 \mid L_{i}\neq L_{j})=b$.
\end{enumerate}

In model~\eqref{eq:net_gen}, the first part on the right is the label effect, and the second part is a logistic model for the nodal effect. We set $d_Z^0 = 1$, $\bm{F} = (1,1,0,\cdots,0)^{\top}\in \mathbb{R}^{d_Z}$, $d_Z=50$, $a=0.8$, $b=0.1a$, and $c_{\textrm{coef}}=3$. This gives us the simulated network $G$ and covariates $\{\mathbf{Z}^{(i)}\}_{i=1}^{n}$. Next, we generate the DAG $G_{X}$ of $\mathbf{X}^{(i)}$. We set the true order of $G_{X}$ as $\pi^{\star}=(1,2,\cdots,d_X)$, with $(j,j+1) \in E_X$ for $j=0,\ldots,d_X-1$. For each $3 \leq j \leq d_X$, we randomly choose $l$ from $\{1,\ldots,j-2\}$ and set $(l,j) \in E_X$. Thus, except for nodes $j=1,2$, each node has two parents.
We then set $w_{lj} ( h^{\star}_G ( \mathbf{Z}^{(i)} ))$ in~\eqref{eq:GLM-parent} by
\begin{align*}
w_{lj} ( h^{\star}_G ( \mathbf{Z}^{(i)} )) =
\begin{cases}
0 & \text{ if } (l,j) \notin E_X \\
\text{uniformly random from } [-1.0, -0.5]  & \text{ if } (l,j) \in E_X \text{ and } L_i = 1 \\
\text{uniformly random from } [0.5, 1.0]  & \text{ if } (l,j) \in E_X \text{ and } L_i = 2
\end{cases}
\end{align*}
for all $(j,l) \in V_X \times V_X$ and $i \in [n]$. Furthermore, we define the total number of trials in the Binomial distribution to be $T=4$.


\begin{figure}[t!]
\centering
\includegraphics[width=1.0\textwidth]{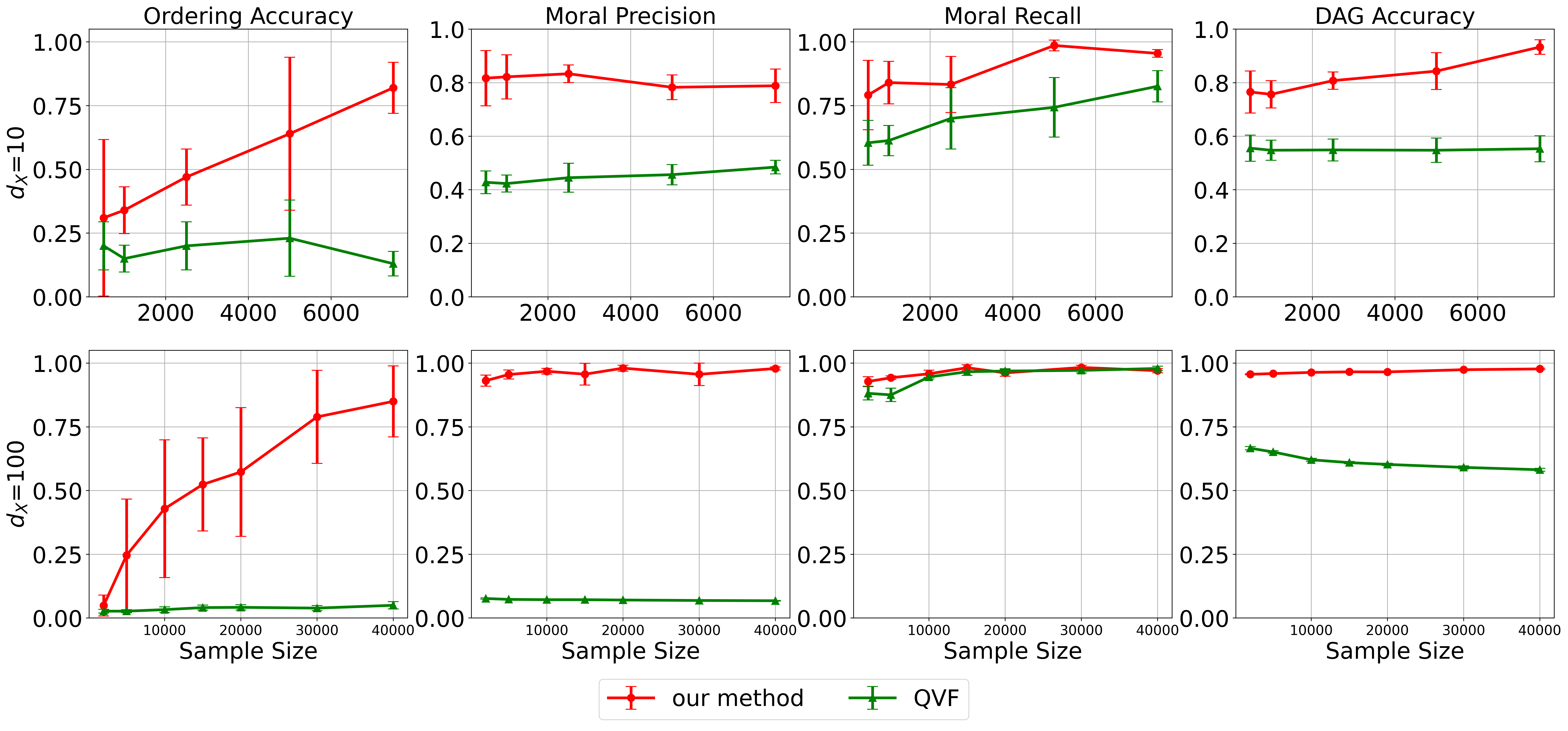}
\caption{Comparison of our algorithm with QVF~\citep{park2017learning} on DAG learning accuracy under linear setup. Both algorithms were run on $10$ independent realizations for each combination of $d_X$ and $n$. The solid dot shows the mean, and the error bar shows one standard deviation across $10$ experiments.}
\label{fig:1}
\end{figure}

We use the linear embedding function from Section~\ref{sec:linear} for node embedding. The estimation procedures follow Sections~\ref{sec:methodolgy} and~\ref{sec:hyperparameter}. The results are in Figure~\ref{fig:1}. We test two DAGs with node sizes $d_X=10$ and $d_X=100$. For $d_X=10$, sample sizes are $n\in\{500,1000,2500,5000,7500\}$; for $d_X=100$, sample sizes are $n\in\{2000,5000,10000,15000,20000,30000,40000\}$. We compare our algorithm with the QVF algorithm in~\cite{park2017learning}, which ignores covariate information.

We use four metrics to evaluate algorithm performance: ordering accuracy, moral precision, moral recall, and DAG accuracy. Ordering accuracy, defined as the Hamming distance between $\hat{\pi}$ and $\pi^{\star}$, measures order recovery accuracy and Step~3 of Algorithm~\ref{alg:ODS-order}. Moral precision and recall are the precision and recall of moral graph estimation $\cup_{j \in V_X} \hat{\mathcal{N}}(j)$ with the ground truth $\cup_{j \in V_X} \mathcal{N}(j)$, measuring performance on moral graph recovery and Step~2 of Algorithm~\ref{alg:ODS-order}. The DAG accuracy is the Hamming distance between $\hat{G}_X$ and $G_X$, measuring the ultimate goal and Algorithm~\ref{alg:ODS-order}.


With heterogeneous DAGs, our method outperforms QVF in all four measurements. While both methods show similar moral recall, QVF has poor moral precision, especially with large node size, indicating low true positive rate when ignoring DAG heterogeneity. Our method achieves consistent ordering and DAG accuracy, approaching $1$ with increasing sample size. In contrast, QVF's ordering accuracy is low, particularly when $d_X$ is large, indicating poor ordering recovery due to ignored heterogeneity.


\subsection{Nonlinear Setup}

The nonlinear data generation process is similar to the linear setup, except for Step~2 in generating the relationship network $G$. Specifically, we generate $d_Z$-dimensional intermediate covariates $\mathbf{C}^{(i)} \sim \mathcal{N}(\mu_{L_i},\Sigma)$, where $\mu_{L_i}$ and $\Sigma$ are defined as in Section~\ref{sec:exp-linear-setup}. We then obtain $\mathbf{Z}^{(i)}$ by $\mathbf{Z}^{(i)}=\sin (\mathbf{C}^{(i)})$, applied element-wise.

\begin{figure}[t!]
\centering
\includegraphics[width=1.0\textwidth]{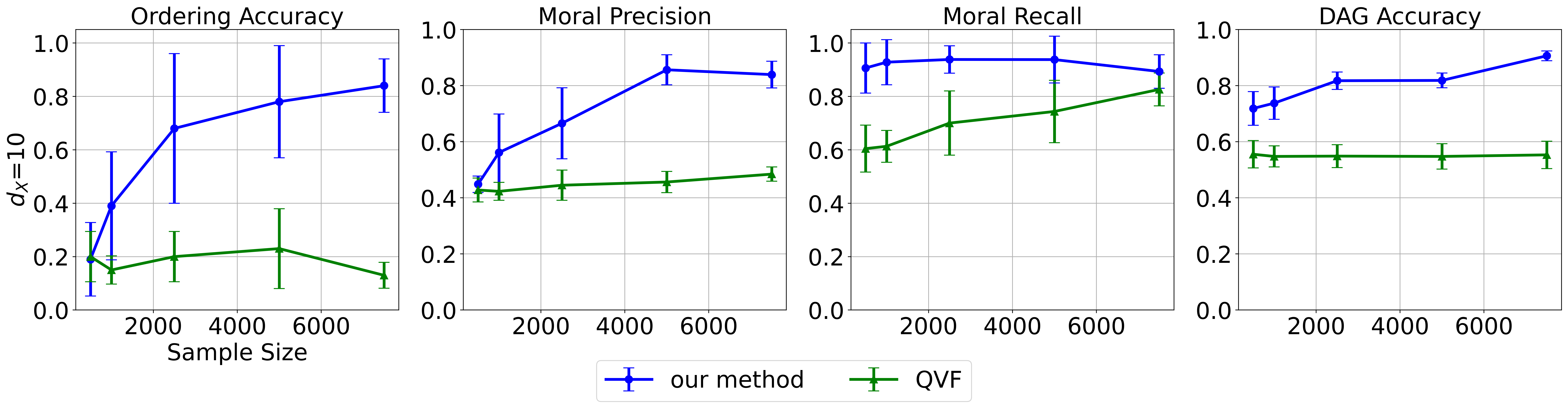}
\caption{Comparison of our algorithm with QVF~\citep{park2017learning} on DAG learning accuracy under nonlinear setup. Both algorithms were run on $10$ independent realizations for each combination of $d_X$ and $n$. The solid dot shows the mean, and the error bar shows one standard deviation across $10$ experiments.}
\label{fig:2}
\end{figure}

We use GAEs from Section~\ref{sec:nonlinear} for node embedding with input feature $\mathbf{Z}^{(i)}$ and set $d_{Z}^{0}=4$. The rest follows Section~\ref{sec:exp-linear-setup}. Due to GAEs' high computational cost for large $n$ and $d_X$, we only experiment with $d_X=10$ and $n\in\{500,1000,2500,5000,7500\}$. Results in Figure~\ref{fig:2} show our method outperforms QVF under nonlinear setup, especially in neighborhood selection and moral precision. Our method achieves consistency in order and DAG structure recovery, unlike QVF.

\section{Real-World Application}
\label{sec:real-data-exp}

In this section, we demonstrate our algorithm's practical usefulness using real-world web visit data from Alipay, which records Chinese users' behavior on multiple Alipay websites during the COVID-19 pandemic. Our goal is to understand user transitions between these websites. For instance, to use public transport, people need to show their green code, a digital code indicating COVID-19 exposure risk. Thus, the public transport payment website should cause visits to the green code website. Our aim is to reveal such relationships in the data, aiding the operational team in designing better strategies for customers.

We collect data $\{\mathbf{X}^{(i)}, \mathbf{Z}^{(i)}\}_{i=1}^{n}$ for $n=6{,}000$ users. The vector $\mathbf{X}^{(i)} \in \{0,1,\ldots,5000\}^{17}$ represents the visit counts of user $i$ to 17 Alipay websites, where the $j$-th entry indicates the number of visits to page $j$ in the past 5,000 records. The vector $\mathbf{Z}^{(i)} \in \mathbb{R}^{21}$ comprises 21 features of user $i$, such as sex and age. The network $G$ captures payment relationships, with an edge between users $i$ and $j$ indicating a money transfer between them on Alipay.

\begin{figure}[t!]
\centering
\includegraphics[width=1.0\textwidth]{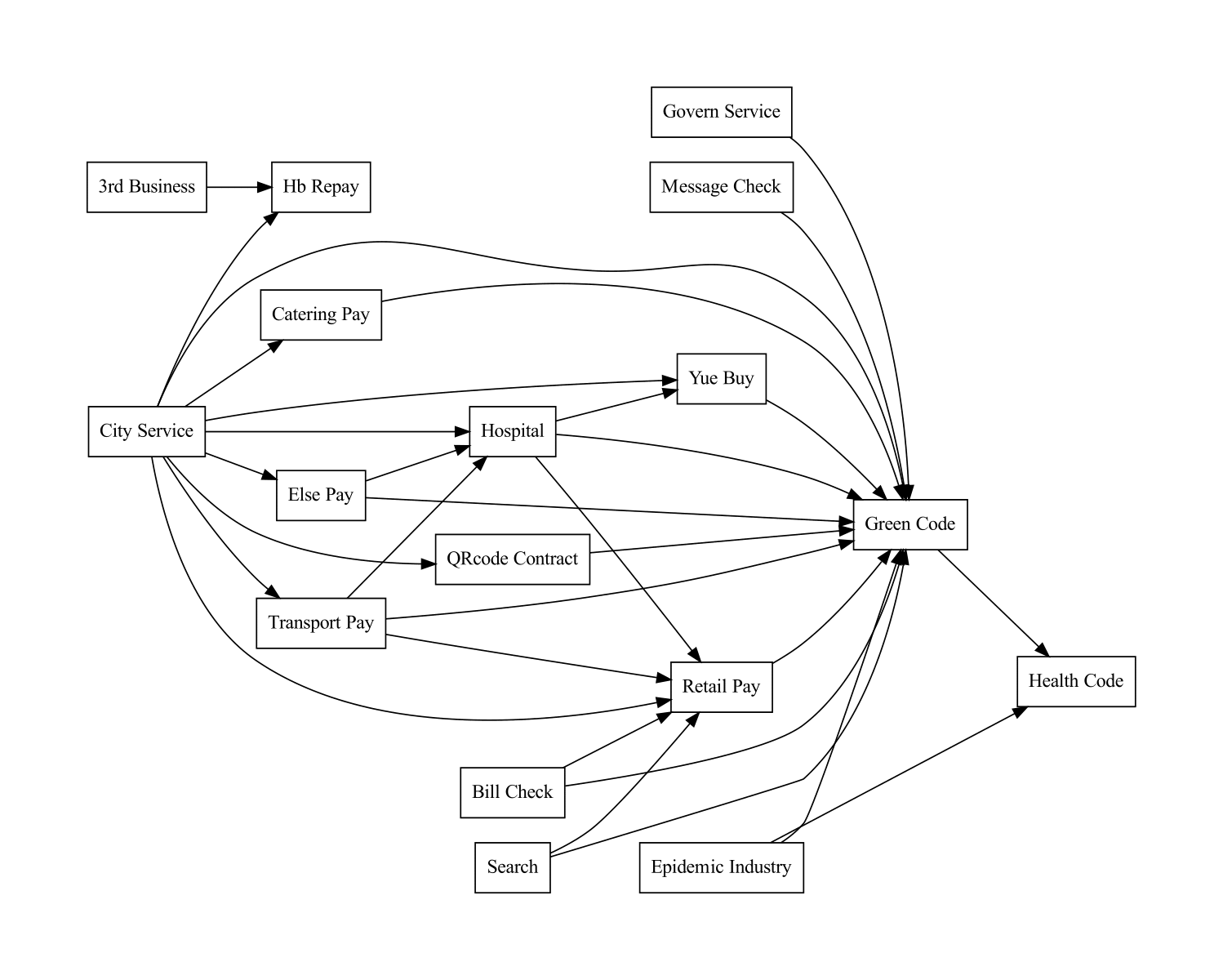}
\caption{DAG estimation result of the $17$ websites of Alipay.}
\label{fig:3}
\end{figure}

\begin{table}[htbp]
\centering
\begin{tabular}{|c|l|}
\hline
\multicolumn{1}{|c|}{\textbf{Term}} & \multicolumn{1}{c|}{\textbf{Explanation}} \\
\hline
3rd Business & Applications operated by the third party service providers\\ \hline
Hb Repay & Huabei repayment \\ \hline
Catering Pay & Payment on catering and restaurant\\ \hline
Retail Pay & Payment on retail and supermarket\\ \hline
Transport Pay & Payment on public transportation like bus and metro\\ \hline
Else Pay & Payment which can not be categorized into the former several types\\ \hline
Hospital & Service provided by the hospital\\ \hline
QRcode Contract & QR payment code used by merchants to receive payment from customers \\ \hline
Bill Check & Bill and check owned by users\\ \hline
Search & Search engine to search for services\\ \hline
Govern Service & The service provided by the government\\ \hline
Message Check & Phone notification to remind users to check messages \\ \hline
Yue Buy & A type of currency fund\\ \hline
Epidemic Industry & Industry related to the COVID-19, such as nucleic acid testing provider\\ \hline
Green Code & Digital code showing the risk of users to COVID-19 exposure \\ \hline
Health Code & Digital code containing the information of user's health condition \\
\hline
\end{tabular}
\caption{The explanations of the abbreviations of the $17$ websites.}
\label{tab:table1}
\end{table}

When implementing Algorithm~\ref{alg:ODS-order}, we use GAEs from Section~\ref{sec:nonlinear} to estimate the node embedding function. The DAG estimation result of the 17 websites is shown in Figure~\ref{fig:3}. Explanations of the abbreviations are summarized in Table~\ref{tab:table1}. The estimated graph reveals interesting causal/directional relationships between different scenes. For instance, "City Service"$\to$"Transport Pay"$\to$"Green Code" demonstrates that users utilize the city service website to find public transportation routes such as subways and buses. Before boarding, they make payments via Alipay and display their green code to verify their health status, aligning with our earlier hypothesis about the link between transportation and green code. Additionally, "City Service"$\to$"Catering Pay"$\to$"Green Code" suggests that users first utilize the city service to find and book restaurants. They then make payments online via Alipay and are required to present their green code to confirm their health status before entering the restaurant. In the same way, "City Service"$\to$"Retail Pay"$\to$"Green Code" indicates that users first utilize the city service to find available markets and make reservations. They subsequently complete their payments online via Alipay, and to receive their orders, they must present their green code to the delivery personnel.
From the perspective of merchants, "City Service"$\to$"QRcode Contract"$\to$"Green Code" indicates that small business owners use city service to apply for QR code payments, with Alipay checking the green code to mitigate health risks. Additionally, "City Service"$\to$"Hospital"$\to$"Yue Pay" shows that users search for nearby hospitals via city service and pay medical bills through Yu'e Bao. "City Service"$\to$"Hospital"$\to$"Green Code" demonstrates that users must present their green code before registering after finding nearby hospitals. Additionally, "Search"$\to$"Retail Pay" indicates that users utilize a search engine to find items on the Tmall market and make payments via Alipay.
Conversely, "Govern Service"$\to$"Green Code" suggests that to access government services, such as obtaining social security funds, users must present their green code before receiving the service in person. Notably, "Epidemic Industry"$\to$"Green Code"/"Health Code" indicates that users are required to show their green code before undergoing nucleic acid testing. Interestingly, "Green Code"$\to$"Health Code" implies that if users check their green code and find it is not green, they will subsequently verify their health code, which might be yellow or red.

As illustrated earlier, the majority of the causal/directional relationships depicted in Figure~\ref{fig:3} can be comprehensively and reasonably explained. Another significant finding from Figure~\ref{fig:3} is that the green code is intricately linked to all facets of people's lives. Given that the green code was essential for nearly all activities during the COVID pandemic in China, this conclusion is intuitively accurate, further confirming the validity of our algorithm.

\section{Conclusion}

In this paper, we introduce the heterogeneous DAGs model with network structured covariates, which are common in practice. We propose an algorithm to estimate the DAG structure. Our algorithm's correctness is demonstrated through simulations and real-world data. In simulations, our method outperforms state-of-the-art DAG discovery algorithms for count data that ignore heterogeneity, showing the importance of considering heterogeneity in causal discovery. We apply our algorithm to Alipay website visit data, providing intuitive results for operational strategies. Future research will explore theoretical guarantees and extend our model beyond binomial data.

\newpage

\bibliography{boxinz-papers}


\newpage

\appendix

\end{document}